\pdfoutput=1

\documentclass[11pt]{article}

\usepackage[]{EMNLP2022}
\usepackage{lipsum}
\usepackage{times}
\usepackage{latexsym}

\usepackage[T1]{fontenc}

\usepackage[utf8]{inputenc}

\usepackage{microtype}

\usepackage{inconsolata}
\usepackage{graphicx}
\usepackage{subfigure}
\usepackage{amsmath}
\usepackage{tabularx}
\usepackage{makecell}
\usepackage{booktabs}
\usepackage{amssymb}
\usepackage{multirow}
\usepackage[ruled,linesnumbered]{algorithm2e}

\title{Improving Complex Knowledge Base Question Answering via Question-to-Action and Question-to-Question Alignment}

\author{Yechun Tang, Xiaoxia Cheng, Weiming Lu\thanks{\hspace{1mm} Corresponding author.} \\
        College of Computer Science and Technology , Zhejiang University \\ \texttt{\{tangyechun, zjucxx, luwm\}@zju.edu.cn}}

\begin{document}
\maketitle
\begin{abstract}
Complex knowledge base question answering can be achieved by converting questions into sequences of predefined actions. However, there is a significant semantic and structural gap between natural language and action sequences, which makes this conversion difficult. In this paper, we introduce an alignment-enhanced complex question answering framework, called ALCQA, which mitigates this gap through question-to-action alignment and question-to-question alignment. We train a question rewriting model to align the question and each action, and utilize a pretrained language model to implicitly align the question and KG artifacts. Moreover, considering that similar questions correspond to similar action sequences, we retrieve top-k similar question-answer pairs at the inference stage through question-to-question alignment and propose a novel reward-guided action sequence selection strategy to select from candidate action sequences. We conduct experiments on CQA and WQSP datasets, and the results show that our approach outperforms state-of-the-art methods and obtains a 9.88\% improvements in the F1 metric on CQA dataset. Our source code is available at \url{
https://github.com/TTTTTTTTy/ALCQA}.
\end{abstract}

\section{Introduction}

Complex knowledge base question answering (CQA) aims to answer various natural language questions with a large-scale knowledge graph. Compared to simple questions with single or multi-hop of relations, complex questions have more kinds of answer types such as \textit{numeric} or \textit{boolean} types and require more kinds of aggregation operations like \textit{min/max} or \textit{intersection/union} to yield answers. Semantic parsing approaches typically map questions to intermediate logical forms such as query graphs ~\citep{yih2015semantic,bao2016constraint,bhutani2019learning,maheshwari2019learning,lan2020query,qin2021improving}, and further transform them into queries like SPARQL query language. Recently, many works ~\citep{liang2017neural,saha2019complex,ansari2019neural,hua2020few,hua2020retrieve,hua2020less} predefine a collection of functions with constrained argument types and represent the intermediate logical form as a sequence of actions that can be generated using a seq2seq model. Sequence-based methods are natural to accomplish more complex operations by simply expanding the function set, thus making some logically complex questions answerable while they're difficult to answer using query graphs.

\begin{figure*}
\centering 
\includegraphics[width=1.0\textwidth]{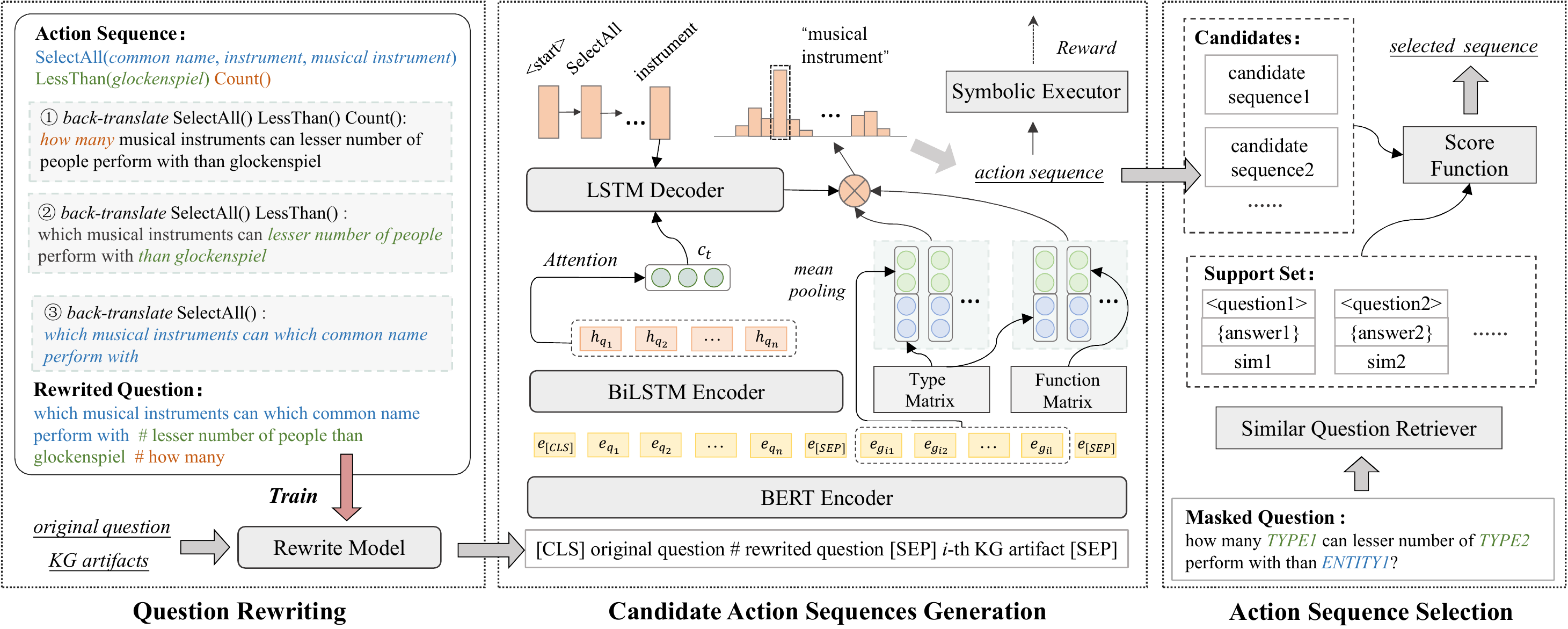} 
\caption{An overview of the proposed approach. The question is first converted into a more structured form, then multiple candidate action sequences are generated by the seq2seq model, and finally the candidate action sequences are scored based on similar question-answer pairs.} 
\label{Fig.main1} 
\end{figure*}

The seq2seq model has been widely used and achieved good results on many text generation tasks, such as machine translation, text summarization and style transfer. In these tasks, the source and the target sequence are both natural language texts and thus share some low-level features. However, semantic parsing aims to transform unstructured texts into structured logical forms, which requires a difficult alignment between them. This problem becomes more serious when the complexity of the question rises. 
Some works propose to solve this problem by modelling the hierarchical structure of logical forms. ~\citet{dong2016language} introduces a sequence-to-tree model with an attention mechanism.
~\citet{dong2018coarse} proposes to decode a sketch of the logical forms which contain a set of functions at first and then decode low-level details like arguments. 
~\citet{guo2021iterative} iteratively segments a span from the question by a segmentation model and parses it using a base parser until the whole query is parsed. 
~\citet{li2021keep} uses a shift-reduce algorithm to obtain token sequences instead of predicting the start and end positions of the span. However, most of these works require intermediate logical forms or sub-questions to train models, which are usually difficult to obtain. ~\citet{guo2021iterative} and ~\citet{li2021keep} propose to pretrain a base parser firstly, and then search good segments that predicted sub logical forms are part of or can be composed into the golden meaning representation. They don't necessarily require training pairs but have the limitation that decomposed utterances are continuous segments of the original question.

In this paper, we propose a novel framework to boost the alignment between unstructured text and structured logical forms. 
We decompose the semantic parsing task into three stages: question rewriting, candidate action sequences generation and action sequence selection. In the question rewriting stage, we utilize a question rewriting model to explicitly transform a query into a set of utterances, each corresponding to a single action, thus reducing the complexity of the question. We propose a two-phase training method to train the rewriting model on the lack of training pairs.
In the candidates generation stage, we build a seq2seq model to generate logical forms with beam search algorithm and consider KG artifacts like entities as candidate vocabularies in the decoding stage. To further align the question and action sequence, we concatenate a question and a KG artifact as input and encode it using a pretrained language model (PLM) like BERT ~\citep{devlin2018bert}. The cross attention mechanism of PLM can effectively align between the question and KG artifacts implicitly, which makes decoding easier. 
Moreover, we innovatively propose to improve complex knowledge base question answering via question-to-question alignment. Motivated by the phenomenon that the more similar two questions are, the more similar their corresponding action sequences will be, we build a memory consisting of question-answer pairs and retrieve a set of question-answer pairs as the support set based on the similarity with the current question during action sequence selection phase. 
We then propose a reward-guided selection strategy that scores each candidate action sequence according to the support set.

Our main contributions are as follow:
\begin{itemize}
\item We propose a novel framework that mitigates the gap between natural language questions and structural logical forms through question-to-action alignment and question-to-question alignment.
\item We propose a novel question rewriting mechanism that rewrites a question into a more structured form without requiring a dataset or adding any constraints, and employ a reward-guided action sequence selection strategy that utilizes similar question-answer pairs to score candidate action sequences.
\item We conduct experiments on several datasets, and experimental results show that our approach is comparable to the state-of-the-art on WQSP dataset and obtains a 9.88\% improvements in the F1 metric on CQA dataset.
\end{itemize}

\section{Methodology}

\subsection{Overview}
In this task, with training set $\mathcal{T} = \{(q_1, a_1), ... , (q_s, a_s)\}$, where $(q_i, a_i)$ is a question-answer pair, the objective is to transform complex questions into logical forms, which can be further derived into KG queries to find answers. We define the logical form as a sequence of actions involving a function and multiple arguments. Following NS-CQA ~\citep{hua2020less}, we design 16 functions with arguments comprised of numerical values and KG artifacts including entities, relations, and entity types. We recognize these arguments in the preprocessing step.
Denote the input question as $q$, the set of predefined functions as function set $\mathcal{F}$, question related numerical values and KG artifacts as argument set $\mathcal{G}$, parameters of model as $\boldsymbol{\theta}$, our goal can be normalized as maximizing the probability $P(\mathcal{L}\mid\boldsymbol{q}\,;\boldsymbol{\theta})$, where $\mathcal{L}$ is the action sequence that produces correct answers and each 
word in $\mathcal{L}$ belongs to $\mathcal{F}$ or $\mathcal{G}$.

As shown in Figure \ref{Fig.main1}, our framework consists of three stages: question rewriting, candidate action sequences generation, and action sequence selection. In the first stage, we rewrite a complex query into a more structured form by a seq2seq model, the details of training the model will be described in \ref{sec:rewriting}. The rewritten query then can be combined with the original question as input, and a newly seq2seq model is used to generate multiple candidate action sequences sequentially. And finally, We retrieve k question-answer pairs that are most similar to the current question from a pre-constructed memory. The candidates are then modified according to the KG artifacts in these k questions and scored based on the comparison results between the execution results and respective answers, separately.

\renewcommand{\algorithmcfname}{Module}
\begin{algorithm}[]
\caption{Question Rewriting Training}
\label{alg:rw}
\KwIn{$\mathcal{T} = \{(q_1, a_1), ... , (q_n, a_n)\}$}
\KwOut{$M_{r}$, which is the trained model for rewriting questions}
Search pseudo action sequences and obtain {$\mathcal{T'} = \{(q_1, a_1, \mathcal{L}_1), ... , (q_n, a_n, \mathcal{L}_n)\}$}, where $\mathcal{L}_i = \{f_1;f_2;...f_k\}$\ is the pseudo action sequence of $(q_i, a_i)$\;
 Train $M_{q}$ which transforms action sequences into questions using $\mathcal{T'}$\;
$\mathcal{Q} \leftarrow \{\}$\;1e-4
\For{$(q_i, \mathcal{L}_i)$ in $\mathcal{T'}$}{
$q_{ori} \leftarrow q_i$ \;
\For{$j \in [k,1]$}{
$\mathcal{L'} \leftarrow \{f_1;f_2;...;f_{j-1}\}$\;
$q_{del} \leftarrow\operatorname{Translate}(\mathcal{L}', M_{q} )$\;
$q'_{ij} \leftarrow\operatorname{Compare}(q_{ori}, q_{del} )$\;
$q_{ori} \leftarrow q_{del}$ \;
}
$\mathcal{Q} \leftarrow \mathcal{Q} \cup \{q_i, \{q'_{i1};q'_{i2};...;q'_{ik}\}\}$ \;
}
Train $M_{r}$ using $\mathcal{Q}$\;
\end{algorithm}

\subsection{Question Rewriting\label{sec:rewriting}}
An action sequence consists of multiple consecutive actions, and it is difficult for the seq2seq model to decide which part of the question to focus on when generating each action. 
We train a question rewriting model that transform a query into a set of utterances which are concatenated by the symbol "\#" and each utterance corresponds to a single action. With the rewritten question, the model can focus on a certain part of the question when generating action in the sequence, thus reducing the difficulty of decoding.

To train the rewriting model, we require an adequate training corpus which is difficult to obtain. 
On the lack of golden datasets, we propose a two-phase approach to convert queries into rewritten questions and use them for training of the rewriting model as shown in Module \ref{alg:rw}. In the first phase (line 1-2), we employ a breadth-first search algorithm to find pseudo action sequences for some questions, and then train a seq2seq model that translates an action sequence into a query. In the second phase (line 3-13), we construct a training corpus for question rewriting based on searched question-logical form pairs and the model trained in the previous stage. 
Specifically, given an action sequence $\mathcal{L}= \{f_1;f_2;...f_k\}$, we delete the last action $f_k$, back-translate the shorter action sequence into a new query, and compare it with the original question. We can determine that the tokens which appear in the original question but not in the current generated question are the ones we should most focus on when generating the deleted action. For example, the left part of Figure \ref{Fig.main1} illustrates the process of decomposing the question "\textit{how many musical instruments can lesser number of people perform with than glockenspiel}". 
We firstly delete the last action "\textit{Count}()" and then the seq2seq model translates the newly formed sequence "\textit{SelectAll}(...)\textit{LessThan}(...)" into query "\textit{which musical instruments can lesser number of people perform with than glockenspiel}". The words "\textit{how many}" should be paid more attention because they do not appear in the generated question. We iteratively perform \textit{delete}, \textit{back-translate} and \textit{compare} operations until the action sequence is empty and concatenate the compare results of each step using symbol "\#". 

Thus, we can construct the question rewriting dataset $\mathcal{Q}$ and train a question rewriting model $M_r$. To make the rewriting model learn to output KG artifacts in the rewritten query, we concatenate the original question and KG artifacts as input, and wrap KG artifacts with symbols like  $\langle  \rm{entity}\rangle$  and  $\langle\rm{/entity}\rangle$. We initialize models in both phases using BART ~\citep{lewis2020bart}, an outstanding pretrained seq2seq model that demonstrates high performance on a wide range of generation tasks, and finetune them by constructed datasets.

\subsection{Encoder-decoder Architecture}

We use BERT and BiLSTM ~\citep{hochreiter1997long} to construct the encoder. Given a question $q$ with $n$ tokens and the argument set $\mathcal{G}=\{g_{1},..., g_{m}\}$, where $m$ is the size of argument set with respect to $q$ and $g_{i}=\{g_{i1},..., g_{il}\}$ is a KG artifact or numerical value with $l$ tokens, we concatenate the question and each argument separately using [SEP] as the delimiter to construct BERT input sequences. In this case, we obtain question embedding $E_{q} \in \mathbb{R}^{n \times d_{e}}$, and argument embedding $\vec{g}_{i} \in \mathbb{R}^{d_{e}}$ by mean pooling over $E_{g_{i}}$. 
We then stack embeddings of arguments to construct a matrix $E_{\mathcal{G}} \in \mathbb{R}^{m \times d_{e}}$  and feed $E_{q}$ into a BiLSTM encoder to obtain the final question representation $H \in \mathbb{R}^{n \times d_{h}}$.
\begin{align}
    E &= {\rm BERT_{}}(\{{\rm [CLS]},q, {\rm [SEP]},g_{i}, {\rm [SEP] }\}) \nonumber \\
    H &= {\rm BiLSTM}(E_{q}) \nonumber \\
    \vec {g}_{i} &= {\rm MeanPooling}(E_{g_{i}})
\end{align}

Decoding is implemented using LSTM, and at each time step, the current hidden state $s_t \in \mathbb{R}^{d_s}$ is updated based on the hidden state and output of the previous time step as follows:
\begin{align}
    s_t &= {\rm LSTM}([o_{t-1};\tau_{t-1};c_{t}], s_{t-1}) \nonumber \\
    c_{t} &= \sum_{i} \alpha_{ti} h_i  \nonumber \\ 
    \alpha_{t} & = {\rm Softmax}(e_{t}) \nonumber \\ 
    e_{t} &= s_{t-1} W_a H^T 
\end{align}

\noindent where [;] denotes vector concatenation. $o_{t-1}$ is the embedding of output in the last step which obtains from learnable embedding matrix $W_{func}$ if the output is a function or from $E_{\mathcal{G}}$ if the output is an argument. $\tau_{t-1}$ is a vector that obtains from learnable embedding matrix $W_{type}$ according to the type of last output.
$c_{t}$ is the context vector resulting from the weighted summation of $h_i$, the \textit{i}-th row of the question embedding $H$, based on the attention mechanism. $W_a  \in \mathbb{R}^{d_{s} \times d_{h}}$ is a projection matrix.



We then calculate the vocabulary distribution based on hidden state $s_t$. Our vocabulary consists of two parts, a fixed vocabulary containing a collection of predefined functions and a dynamic vocabulary consisting of arguments, i.e., numerical values and KG artifacts related to the question. We feed $s_t$ through one linear layer $W_o$ and a softmax function to compute the probability of each word in the fixed vocabulary. To obtain the probabilities of the words in the dynamic vocabulary, we project the hidden vectors $s_t$ to the same dimension through the projection matrix $W_p \in \mathbb{R}^{d_{s} \times d_{e}}$ and then compute the similarities with each word by taking the dot product.
\begin{align}\label{eq.fix}
    P_{fix} &= {\rm Softmax}(W_{o}s_{t}) \nonumber \\
    P_{dyn} &= {\rm Softmax}(s_{t}W_{p}E_{\mathcal{G}}^T)
\end{align}

Next, we calculate the probability $P_t$ that generate from the fixed vocabulary at the current time step through a linear layer followed by the activation function, and combine the two vocabulary distributions based on $P_t$. 
Note that if $w$ is a word in fixed vocabulary, then $P_{dyn}(w)$ is zero; similarly $P_{fix}(w)$ is zero when $w$ is in dynamic vocabulary.
\begin{align}
    P(w) &= P_{t}P_{fix}(w) + (1-P_{t})P_{dyn}(w)  \nonumber \\
    P_{t} &= \sigma(W_{f}c_{t})
\end{align}

\subsection{Reward-guided Action Sequence Selection Strategy}
To improve accuracy, we generate multiple candidate action sequences with beam search algorithm and design a reward-guided action sequence selection strategy. In general, the more similar the structure and semantics of the two questions are, the more similar their corresponding action sequences will be. Therefore, we propose that similar questions can be used to help the selection of correct action sequence. Specifically, we build a memory consisting of question-answer pairs in the training set. Note that we don't require golden logical forms of these questions. 

To retrieve similar questions with answers from memory, we use edit distance to calculate the similarity between two questions. To improve the generalization of the questions, we replace the entity mentions, type mentions and numerical values in the questions with the symbol [ENTITY], [TYPE] and [CONSTANT], respectively. We don't mask relations because it is always hard to recognize relation mentions. In addition, the presence of some antonyms including \textit{atmost} and \textit{atleast}, \textit{less} and \textit{greater}, can lead to the exact opposite semantics of questions with similar contexts. Therefore, we construct a set of antonym pairs and set the similarity to 0 when there is an antonym pair in the two questions. We retrieve $k$ question-answer pairs with the highest similarity to form the support set $S = \{\{q_1, a_1, d_1\}, ..., \{q_k, a_k, d_k\}\}$, where $d_i$ is the similarity computed by edit distance.

\begin{figure}[htbp]
\centering

\subfigure[question-to-question alignment]{
\begin{minipage}[t]{1.0\linewidth}
\centering
\includegraphics[width=1.0\linewidth]{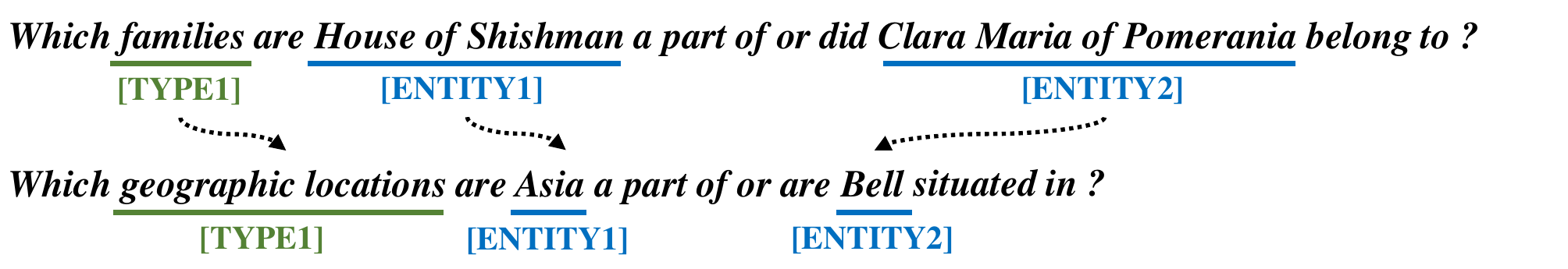}
\end{minipage}%
}%
\quad
\subfigure[action-to-action alignment]{
\begin{minipage}[t]{1.0\linewidth}
\centering
\includegraphics[width=1.0\linewidth]{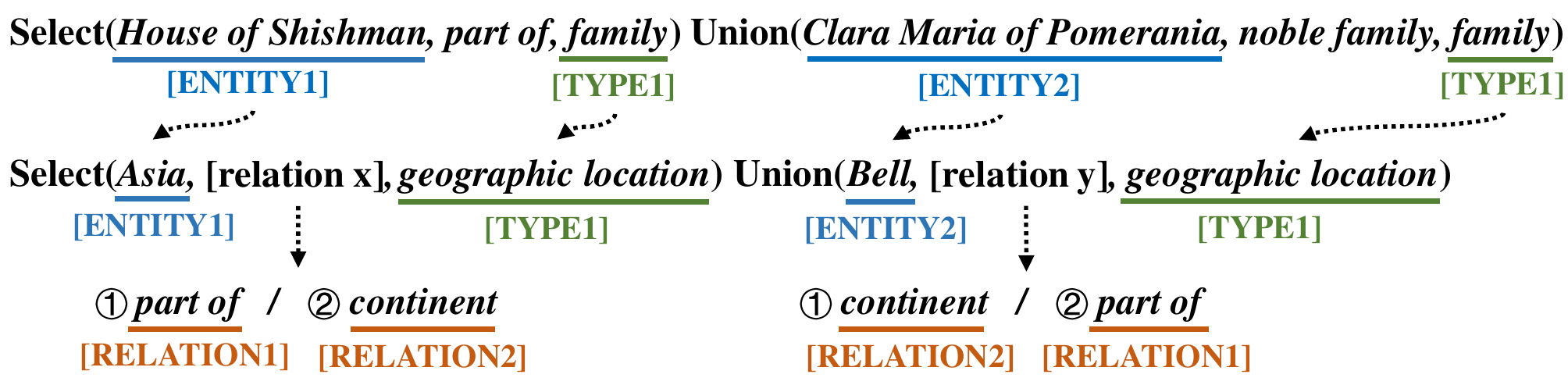}
\end{minipage}
}%
\centering
\caption{An example of adjusting candidate action sequences. The upper and lower parts of (a) are the original question and a question in the support set, respectively. We first obtain a relation-masked action sequence (the second line of (b)) based on the alignment results of entities and types between two questions as shown in (a), and then output multiple action sequences according to all possible combinations of relations.}
\label{Fig.main3}
\end{figure}

We then propose a reward-guided action sequence selection strategy that scores each candidate action sequence according to its fitness to the retrieved support set. Specifically, given a candidate $\mathcal{A}_{i}$ and an item $\{q_j, a_j, d_j\}$ in the support set, we adjust the arguments in $\mathcal{A}_{i}$ to arguments of ${q_j}$ according to their positions in the text as Figure \ref{Fig.main3}, and then score it by compute F1 scores between $a_j$ and execution results of modified sequences on the lack of golden action sequences. Due to the positions of the relations being unknown, we obtain all possible orders of relations and generate multiple modified action sequences. We  then take the highest F1 as the score of $\{q_j, a_j, d_j\}$ to $\mathcal{A}_{i}$ and denote it as $r_{i}^{j}$. The overall score of $\mathcal{A}_{i}$ then can be calculated as follows:
\begin{align}
    s_{i}&=\frac{\sum_{j=1}^{k} d_{j} r_{i}^{j} }{\sum_{j=1}^{k} d_{j}}
\end{align}
where $\sum_{j=1}^{k} d_{j}$ is a normalized term. 
We take the candidate action sequence with the highest score as the output
sequence in the inference stage.

\subsection{Training}
We use REINFORCE~\citep{williams1992simple} algorithm to train our model. We view F1 scores of the answers generated by predicted action sequence with respect to ground-truth answers as original rewards. To improve the stability of training, we use the adaptive reward function ~\citep{hua2020less} to adjust rewards. 
Moreover, we use a breadth-first search algorithm on a subset of data to obtain pseudo-action sequences 
and pretrain the model to prevent the cold start problem. 

\section{Experiments}

\subsection{Experimental Setup}
Our method aims to solve various complex questions, and we mainly evaluate it on ComplexQuestionAnswering  (CQA) ~\citep{saha2018complex} dataset which is a large-scale KBQA dataset containing seven types of complex questions, as shown in Table \ref{Table.exp1}. We show the details and some examples of this dataset in Appendix \ref{sec:cqa}.  We also conduct experiments on WebQuestionsSP  (WQSP) ~\citep{yih2015semantic} which contains 4737 simple questions. The results show that our method also works well on simple datasets. 

We employ standard F1-measure between predicted entity set and ground truth answers as evaluation metrics. For some categories whose answers are boolean values or numbers on CQA dataset, we view answers as single-value sets and compute the corresponding F1 scores. The training details and model parameters can be found in Appendix \ref{sec:details}

\begin{table*}
\centering
\small
\begin{tabular}{lcccccccc}
\toprule
\textbf{Question Category} & \textbf{KVmem} & \textbf{CIP-All} & \textbf{CIP-Sep}  & \textbf{NSM} & \textbf{MRL-CQA} & \textbf{MARL} & \textbf{NS-CQA} & \textbf{Ours} \\ \midrule
Simple Question  & 41.40\% & 41.62\% & \textbf{94.89\%} & 88.33\% & 88.37\% & 88.06\% & \underline{88.83\%} & 88.73\% \\ 
Logical Reasoning & 37.56\% & 21.31\% & \underline{85.33\%} & 81.20\% & 80.27\% & 79.43\% & 81.23\% & \textbf{88.73\%} \\ 
Quantitative Reasoning & 0.89\% & 5.65\% & 33.27\% & 41.89\% & 45.06\% & 49.93\% & \underline{56.28\%} & \textbf{76.30\%} \\ 
Comparative Reasoning & 1.63\% & 1.67\% & 9.60\% & 64.06\% & 62.09\% & 64.10\% & \underline{65.87\%} & \textbf{83.09\%} \\ \midrule
Verification (Boolean)  & 27.28\% & 30.86\% & 61.39\% & 60.38\% & 85.62\% & \underline{85.83\%} & 84.66\% & \textbf{88.18\%} \\ 
Quantitative (Count) & 17.80\% & 37.23\% & 48.40\% & 61.84\% & 62.00\% & 60.89\% & \underline{76.96\%} & \textbf{80.41\%} \\ 
Comparative (Count)  & 9.60\% & 0.36\% & 0.99\% & 39.00\% & 40.33\% & 40.50\% & \underline{43.25\%} & \textbf{60.80\%} \\ \midrule
Overall macro F1 & 19.45\% & 19.82\% & 47.70\% & 62.39\% & 66.25\% & 66.96\% & \underline{71.01\%} & \textbf{80.89\%} \\
Overall micro F1 & 31.18\% & 31.52\% & 73.31\% & 76.01\% & 77.71\% & 77.71\% & \underline{80.80\%} & \textbf{85.31\%} \\ \bottomrule
\end{tabular}
\caption{The overall performances on CQA dataset. Best results are bolded for each category and second-best results are underlined.}
\label{Table.exp1}
\end{table*}

\subsection{Baselines}
We compare our framework with seq2seq based methods. KVmem~\citep{saha2018complex} presents a model consisting of a hierarchical encoder and a key value memory network. 
CIPITR~\citep{saha2019complex} proposes to mitigate reward sparsity with auxiliary rewards and restricts the program space to semantically correct programs.
CIPITR proposes two training ways, one training a single model for all question categories, denoted by CIP-ALL, and the other training a separate model for each category, denoted by CIP-SEP. 
NSM~\citep{liang2017neural} utilizes a key-variable memory to handle compositionality and helps find good programs by pruning the search space. 
MRL-CQA~\citep{hua2020few} and MARL~\citep{hua2020retrieve} propose meta-reinforcement learning approaches that effectively adapts the meta-learned programmer to new questions to tackle potential distributional biases, where the former uses an unsupervised retrieval model and the latter learns it alternately with the programmer from weak supervision. 
NS-CQA~\citep{hua2020less} presents a memory buffer that stores high-reward programs and proposes an adaptive reward function to improve training performance. 
SSRP~\citep{ansari2019neural} presents a noise-resilient model that is distant-supervised by the final answer. 
CBR-KBQA~\citep{das2021case} generates complex logical forms conditioned on similar retrieved questions and their logical forms to generalize to unseen relations.

We also compare our method with graph-based methods on WQSP dataset. STAGG~\citep{yih2015semantic} proposes a staged query graph generation framework and leverages the knowledge base in an early stage to prune the search space. TEXTRAY~\citep{bhutani2019learning} answers complex questions using a novel decompose-execute-join approach. QGG~\citep{lan2020query} modifies STAGG with more flexible ways to handle constraints and multi-hop relations. 
OQGG~\citep{qin2021improving} starts with the entire knowledge base and gradually shrinks it to the desired query graph.


\subsection{Overall Performances}
The overall performances of our proposed framework against KBQA baselines are shown in Table~\ref{Table.exp1} and~\ref{Table.wqsp}. Our framework significantly outperforms the state-of-the-art model on CQA dataset while staying competitive on WQSP dataset. On CQA dataset, our method achieves the best overall performance of 80.89\% and 85.31\% in macro and micro F1 with 9.88\% and 4.51\% improvement, respectively. Moreover, it can be observed that our method achieves the best result on six of seven question categories. On \textit{Logical Reasoning} and \textit{Verification (Boolean)}, which are relatively simpler, our model obtain a 3.40\% and 2.35\% improvement in macro F1, respectively. On \textit{Quantitative Reasoning}, \textit{Comparative Reasoning}, \textit{Quantitative (Count)} and \textit{Comparative (Count)}, whose questions are complex and hard to parse, out model obtain a considerable improvement. To be specific, the macro F1 scores increase by 20.02\%, 17.22\%, 3.45\% and 17.55\%, respectively. Our proposed method doesn't outperform CIP-Sep on \textit{Simple Question} which trains a separate model on this category but still achieves a comparable result with the second-best baseline. On WQSP dataset, our method outperforms all the sequence-based methods and stay competitive with the graph-based method which having the best results. Our method doesn't gain a lot because most questions in this dataset are one hop and simple enough while our frameword aims to deal with various question categories. We don't compare with graph-based methods on CQA dataset because they always start from a topic entity and interact with KG to add relations into query graphs
step by step, which can not solve most question types like \textit{Quantitative Reasoning} and \textit{Comparative Reasoning} in this dataset.

\begin{table}[!ht]
    \centering
    \small
    \begin{tabular}{p{34mm}p{8mm}<{\centering}}
    \toprule
        \textbf{Method} & \textbf{F1}  \\ \midrule
        NSM & 69.0\%  \\
        SSRP & 72.6\% \\
        NS-CQA & 72.0\% \\
        CBR-KBQA\dag & 72.8\% \\ \midrule
        STAGG & 66.8\% \\
        TEXTRAY & 60.3\% \\
        QGG & \textbf{74.0\%}\\
        OQGG & 66.0\%\\ \midrule
        Ours &  \underline{73.6\%} \\ 
        \bottomrule
    \end{tabular}
\caption{The overall performances on WQSP dataset. \dag \quad denotes supervised training.}
\label{Table.wqsp}
\end{table}

The experimental results demonstrate the ability of our method to parse complex questions and generate correct action sequences. The main improvement of the proposed method comes from two aspects. On the one hand, we employ a rewrite model to decompose a complex question into several utterances, allowing the decoder to focus on a shorter part when decoding each action. On the other hand, we make full use of existing question-answer pairs and determine the structure of action sequences indirectly through the alignment between question-question pairs.

\subsection{Ablation Studies}


We conduct a series of ablation studies on CQA dataset to demonstrate the effectiveness of the main modules in our framework. To explore the impact of question rewriting module, we remove it and only use the original question as input of the seq2seq model. The performance drops by 1.79\% in macro F1 as shown in Table ~\ref{Table.exp2}. To prove the effectiveness of the action sequence selection module, we generate candidate action sequences using beam search mechanism and directly use the action sequence with the highest probability as the output instead of selecting by action sequence selection module. The macro F1 drops by 2.49\% after removing this module. To verify that the cross-attention mechanism in BERT can lead to alignment between question and KG artifacts and further improve the generation result, we encode question and KG artifacts separately and find the performance drops by 0.97\%. Experimental results show that every main module in our framework has an important role in performance improvement.

\begin{table}[!ht]
    \centering
    \small
    \begin{tabular}{lcc}
    \toprule
        \textbf{Settings } & \textbf{macro F1} & \textbf{micro F1} \\ \midrule
         Full Model  & 80.89\% &85.31\%\\ \midrule
        w/o question rewriting & 79.10\% & 84.15\%\\ 
        w/o candidates selection  & 78.40\% & 83.55\%\\
        w/o cross-attention & 79.92\% & 84.63\%\\ 
        \bottomrule
    \end{tabular}
\caption{Ablation studies on main components.}
\label{Table.exp2}
\end{table}

To explore the impact of employing different underlying embeddings, we conduct experiments on two settings, initializing an embedding matrix randomly and encoding with BERT. We finetune the embedding matrix during the training stage in the first setting while freezing the parameters of the BERT model. As shown in Table \ref{Table.exp3}, BERT embedding achieves the best result and improves by 4.40\% compared to random embedding. It is reasonable because BERT is pretrained with a large corpus to represent rich semantics and uses a cross-attention mechanism to align the question and KG artifacts better. Note that our proposed method still outperforms state-of-the-art methods without using BERT.

\begin{table}[!ht]
    \centering
    \small
    \begin{tabular}{lcc}
    \toprule
        \textbf{Settings} & \textbf{macro F1}  & \textbf{micro F1}\\ \midrule
        Random Embedding & 76.49\% & 81.63\%\\ 
        BERT Embedding  & 80.89\% & 85.31\%\\  \bottomrule
    \end{tabular}
\caption{Ablation studies for different underlying embeddings.}
\label{Table.exp3}
\end{table}



\begin{figure}
\centering 
\includegraphics[width=1.0\linewidth]{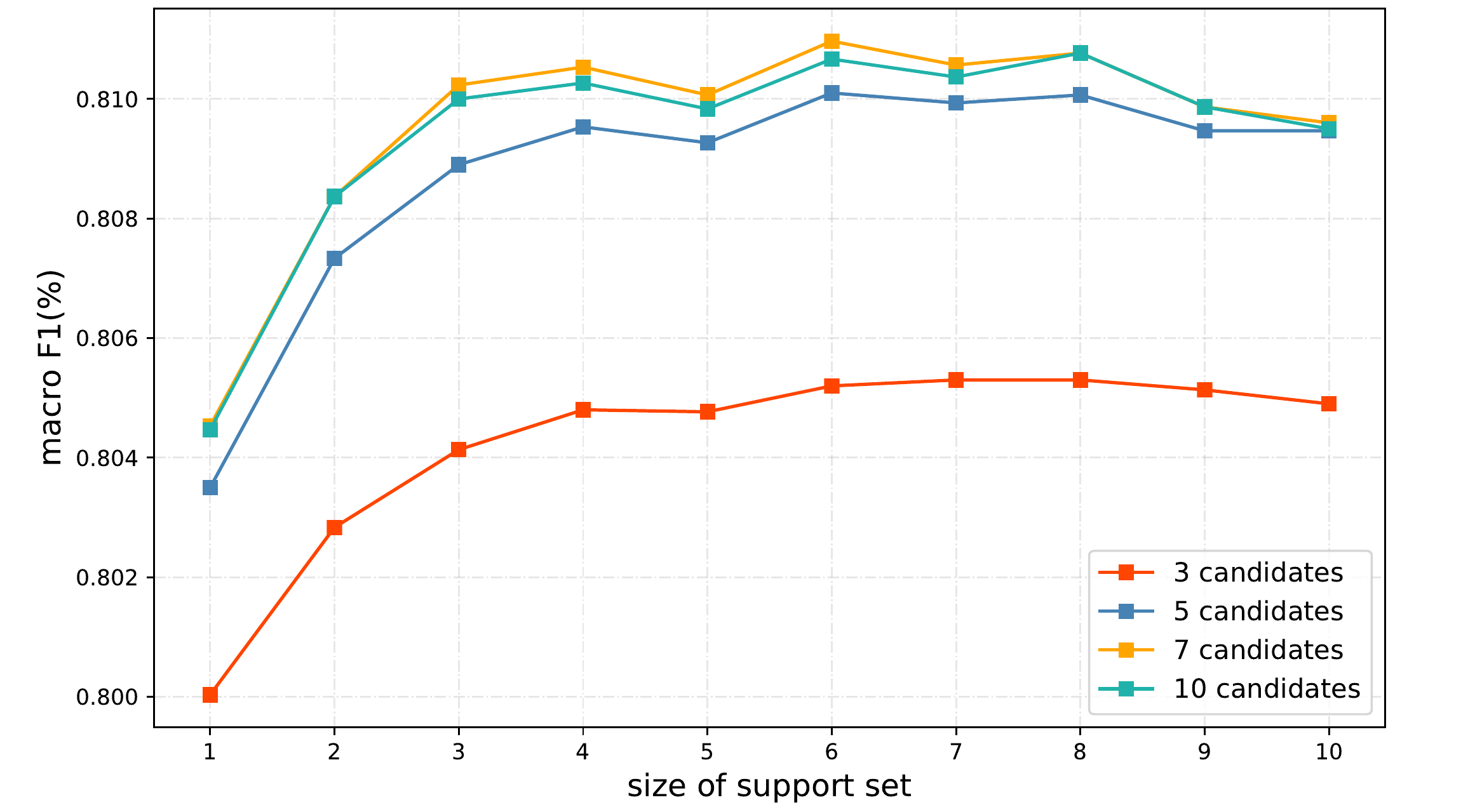} 
\caption{Trends of macro F1 when the size of support set increases.} 
\label{Fig.data} 
\end{figure}

To investigate the effect of the number of candidate action sequences and the size of the support set on the selection of action sequences, we conduct experiments and plot the results in Figure \ref{Fig.data}. 
It can be observed that the macro F1 score increases with the size of the support set at the beginning, whatever the number of candidates.
This trend slows down gradually and the macro F1 score peaks when the size is about 6. Then, as the size of the support set continues to increase, the macro F1 score decreases slightly. It's mainly caused by the simple and rough method we use to calculate the question similarity, which leads to the assumption that similar questions have similar action sequence structures not always hold. In contrast, a certain number of similar questions can alleviate this problem and improve performance. However, when the number reaches a certain level, the newly added questions become less similar to the original questions and introduce noise instead.
In addition, the increase in the number of generated candidates also improves performance. If the number is too high, this boost becomes less apparent or even negative because of the lower quality of the newly added candidates.

\subsection{Case study}

We show some examples to illustrate the ability of our modules. Table \ref{Table.case1}
shows a complex question of category \textit{Quantitative (Count)}. We can observe that the model wrongly predicts the third action in the absence of rewriting module but makes a correct generation with the help of rewritten utterances. It's reasonable because the seq2seq model learns to focus on "approximately 5 people" when predicting the third action. Table \ref{Table.case2} shows a query of category \textit{Verification (Boolean)}. It's confusing for the model to decide which entity to output, and the correct action sequence is given a lower probability. However, it's much easier to choose through action sequence selection module. 
The wrong logical form produces an incorrect result in the majority of cases and thus receives a lower selection score, as shown.

\begin{table}
  \centering
  \small
  \begin{tabular}{p{14mm}|p{54mm}}
  \toprule
  \textbf{Question} & how many works of art feature approximately 5 fictional taxons or people \\
  \midrule
  \textbf{Rewritten Question} & which works of art contain which fictional taxon \# and which common name \# approximately 5 people \# how many    \\\midrule
    \textbf{w/o Module} & SelectAll(\textit{fictional taxon}, \textit{present in work}, \textit{work of art})  SelectAll(\textit{common name}, \textit{present in work}, \textit{work of art})  {\color{red} AtLeast}(\textit{5})  Count() \\ \midrule
    \textbf{w/ Module} & SelectAll(\textit{fictional taxon}, \textit{present in work}, \textit{work of art}) SelectAll(\textit{common name}, \textit{present in work}, \textit{work of art})  {\color{blue} Around}(\textit{5}) Count() \\
  \bottomrule
  \end{tabular}
\caption{Test case on quesiton rewriting module}
\label{Table.case1}
\end{table}

\begin{table}
  \centering
  \small
  \begin{tabular}{p{14mm}|p{54mm}}
  \toprule
  \textbf{Question} & Does Janko Kroner have location of birth at Peraia, Pella and Povazska Bystrica ? \\
  \midrule
    \textbf{w/o Module} & Select(\textit{Janko Kroner}, \textit{place of birth}, \textit{administrative territorial entity}) \# Bool(\textit{\color{red} Povazska Bystrica}) \#  Bool(\textit{Povazska Bystrica}) \newline \textbf{Prob:} 0.6085 \quad \textbf{Selection Score:} 0.7333 \\ \midrule
    \textbf{w/ Module} & Select(\textit{Janko Kroner}, \textit{place of birth}, \textit{administrative territorial entity}) \# Bool(\textit{\color{blue} Peraia, Pella}) \#  Bool(\textit{Povazska Bystrica}) \newline \textbf{Prob:} 0.3519 \quad \textbf{Selection Score:} 1.0000\\
  \bottomrule
  \end{tabular}
\caption{Test case on action sequence selection module}
\label{Table.case2}
\end{table}

\section{Related Work}
Semantic parsing is the task of translating natural language utterances into executable meaning representations. 
Recent semantic parsing based KBQA methods can be categorized as graph-based~\citep{yih2015semantic,bao2016constraint,bhutani2019learning,lan2020query,qin2021improving} and sequence-based~\citep{liang2017neural,saha2019complex,ansari2019neural,hua2020few,hua2020retrieve,hua2020less,das2021case}. Graph-based methods build a query graph which is a graph-like logical form proposed by ~\cite{yih2015semantic}.
~\cite{bao2016constraint} proposed multi-constraint query graph to improve performance.
~\citet{ding2019leveraging} and ~\citet{bhutani2019learning} decomposed complex query graph into a set of simple queries to overcome the long-tail problem.  ~\cite{lan2020query} employed early incorporation of constraints to prune the search space. ~\cite{chen2021formal} leveraged the query structure to constrain the generation of the candidate queries. ~\cite{qin2021improving} generated query graph by shrinking the entire knowledge base. 
Sequence-based methods define a set of functions and utilize a seq2seq model to generate action sequences. ~\citet{liang2017neural} augmented the standard seq2seq model with a key-variable memory to save and reuse intermediate execution results. ~\citet{saha2019complex} mitigated reward sparsity with auxiliary rewards. ~\citet{ansari2019neural} learned program induction with much noise in the query annotation. ~\citet{hua2020few,hua2020retrieve} employed meta-learning to adapt programmer to unseen questions quickly. ~\citet{hua2020less} proposed a adaptive reward function to control the exploration-exploitation trade-off in reinforcement learning.

Compared to graph-based methods, sequence-based methods can generate logical forms directly using the seq2seq model, which is easier to implement and can handle more question categories by simply expanding the set of action functions. However, the semantic and structural gap between natural language utterances and action sequences leads to poor performance on translation.

\section{Conclusion}

In this paper, we propose an alignment-enhanced complex question answering framework, which reduces the semantic and structural gap between question and action sequence by question-to-action and question-to-question alignment. We train a question rewriting model to align question and sub-action sequence in the absence of training data and employ a pretrained language model to align the question and action arguments implicitly. Moreover, we utilize similar questions to help select the correct action sequence from multiple candidates. Experiments show that our framework achieves state-of-the-art on the CQA dataset and performs well on various complex question categories. In the future, how to better align questions with logical forms will be considered.

\section*{Limitations}
In our method, we view KG artifacts as tokens and generate logical forms using a seq2seq model, which can handle more types of complex questions, i.e., superlative quesions without topic entities. However, for single and multi-hop questions, graph-based methods may gain better performance. The reason is that they start from a topic entity and interact with KG to add relations into query graphs step by step, which can prune the search space more effectively. Moreover, we control the vocabulary size through entity and relation recognition, which makes the preprocessing step more complex.

\section*{Acknowledgement}
This work is supported by the National Key Research and Development Project of China (No. 2018AAA0101900),  the Key Research and Development Program of Zhejiang Province, China (No. 2021C01013), CKCEST, and MOE Engineering Research Center of Digital Library.

\bibliography{anthology,custom}
\bibliographystyle{acl_natbib}

\appendix

\begin{table*}[ht]
\centering
\small
\begin{tabular}{lll}
\toprule
\multicolumn{2}{c}{\textbf{Question Category}} & \textbf{Example Question} \\
\midrule
Simple Question (599K) & Direct & Where did the expiration of Brian Hetherston occur ? \\
\midrule
\multirow{3}{30mm}{Logical Reasoning (138K)} & union & Which people were casted in Cab Number 13 or Hearts of Fire ? \\
~ &  Intersection & Who have location of birth at Lourdes and the gender as male ? \\
~ &  Difference & Which people are a native of Grenada but not United Kingdom ? \\
\midrule
Verification (63K) & Boolean & Is United Kingdom headed by Jonas Spelveris and Georgius Sebastos ? \\
\midrule
\multirow{4}{30mm}{Quantitative Reasoning (118K)} & Min/Max & Who had an influence on max number of bands and musical ensembles ? \\
~ &  \multirow{2}*{Atleast/Atmost} & \multirow{2}{95mm}{Which applications are manufactured by atleast 1 business organizations and business enterprises ? }\\
~ & ~ & ~ \\
~ &  exactly/around \textit{n} & Which films had their voice dubbing done by exactly 20 people ? \\
\midrule
\multirow{3}{30mm}{Comparative Reasoning (62K)} & \multirow{3}*{Less/More/Equal} &\multirow{3}{95mm}{Which positions preside the jurisdiction over more number of administrative territories and US administrative territories than Minister for Regional Development ?} \\
~ & ~ & ~ \\
~ & ~ & ~ \\
\midrule
\multirow{6}{30mm}{Quantitative Reasoning (Count) (159K)} & \multirow{2}*{Direct} & \multirow{2}{95mm}{How many nucleic acid sequences encodes Dynein light chain 1, cytoplasmic ?} \\
~ & ~ & ~ \\
~ &  \multirow{2}*{Union} & \multirow{2}{95mm}{How many system software or operating systems are the computing platforms for which Street Fighter IV were specifically designed ? }\\
~ & ~ & ~ \\
~ &  \multirow{2}*{Intersection} & \multirow{2}{95mm}{How many people studied at Harvard University and Ecole nationale superieure des Beaux-Arts ? }\\
~ & ~ & ~ \\
~ &  exactly/around \textit{n} & How many musical instruments are played by atmost 7998 people ? \\
\midrule
\multirow{2}{30mm}{Comparative Reasoning (Count) (63K)} & \multirow{2}*{Less/More/Equal} & \multirow{2}{95mm}{How many administrative territories have less number of cities and mythological Greek characters as their toponym than Bagdad ?} \\
~ & ~ & ~ \\
\bottomrule
\end{tabular}
\caption{The examples of various question types on CQA dataset.}
\label{Table.data}
\end{table*}

\section{CQA Dataset}
\label{sec:cqa}
Complex Question Answering(CQA) dataset contains the subset of the QA pairs from the Complex Sequential Question Answering(CSQA) dataset, where the questions are answerable without needing the previous dialog context. There are 944K, 100K and 156K question-answer pairs in the training, validating and test set, respectively. This dataset has seven types of complex questions, making it difficult for the model to answer correctly. We show some examples of each question category in Table \ref{Table.data}. For simple 
questions, the corresponding action sequence contains only one action, and for some complex questions, the length of action sequence may up to 4.

\section{Training Details}
\label{sec:details}
To compare with previous works and reduce training time, we also randomly select two small subsets(about 1\% each) from the training set to train models. We use BFS algorithm to search pseudo action sequences for the first subset to train the question rewriting model as introduced in \ref{sec:rewriting} and pretrain the action sequence generation model. We use the second one for subsequent reinforcement learning of the action sequence generation model. We evaluate our trained model on the whole test set.

We initialize two models in the question rewriting stage with the base version of BART and finetune them using Adam Optimizer with a learning rate of 1e-5. For the action sequence generation model, we adopt the uncased base version of BERT for underlying embeddings and freeze the parameters to improve training stability. We set the dimension of type embedding to 100, the hidden sizes of one-layer BiLSTM Encoder and LSTM Decoder to 300. We train the model for 100 epochs and 50 epochs using Adam with learning rates of 1e-4 and 1e-5 in the pretraining and reinforcement learning stages, respectively, and finally choose the checkpoint with the highest reward in the development set. We generate 5 candidate action sequences with a beam size of 10, and retrieve 3 questions with a similarity greater than threshold 0.6 as the support set. If no similar question meets the condition, we directly select the top one action sequence generated by beam search as output.

\end{document}